\documentclass[11pt]{article}

\usepackage[final]{acl}

\usepackage{times}
\usepackage{latexsym}

\usepackage[T1]{fontenc}

\usepackage[utf8]{inputenc}

\usepackage{microtype}

\usepackage{inconsolata}

\usepackage{graphicx}

\usepackage{url}
\usepackage{algorithm}
\usepackage{algorithmic}
\usepackage{booktabs}
\usepackage{bbm}
\usepackage{amsmath}
\usepackage{multirow}
\usepackage{xcolor}
\usepackage{colortbl}
\usepackage[table]{xcolor}
\usepackage{array}  
\usepackage{arydshln}   
\usepackage{tcolorbox}
\tcbuselibrary{listings, breakable}
\usepackage{graphicx}
\usepackage{subcaption}
\title{HCRE: LLM-based Hierarchical Classification for Cross-Document Relation Extraction with a Prediction-then-Verification Strategy}

\author{
    Guoqi Ma\textsuperscript{1\thanks{Equal contribution.}}~
    Liang Zhang\textsuperscript{1\footnotemark[1]}~
    Hongyao Tu\textsuperscript{1}~
    Hao Fu\textsuperscript{2}~
    Hui Li\textsuperscript{1}~
    Yujie Lin\textsuperscript{1}\\
\textbf{
    Longyue Wang\textsuperscript{3}~
    Weihua Luo\textsuperscript{3}~
    Jinsong Su\textsuperscript{1\thanks{Corresponding author.}}
}\\
    \textsuperscript{1}School of Informatics, Xiamen University
    \\
    \textsuperscript{2}Li Auto Inc. 
    \\
    \textsuperscript{3}Alibaba International Digital Commerce Group\\
    \texttt{\{guoqima,lzhang\}@stu.xmu.edu.cn, jssu@xmu.edu.cn}
}

\begin{document}
\maketitle

\begin{abstract}
Cross-document relation extraction (RE) aims to identify relations between the head and tail entities located in different documents. 
Existing approaches typically adopt the paradigm of ``\textit{Small Language Model (SLM) + Classifier}''. 
However, the limited language understanding ability of SLMs hinders further improvement of their performance.
In this paper, we conduct a preliminary study to explore the performance of Large Language Models (LLMs) in cross-document RE. 
Despite their extensive parameters, our findings indicate that LLMs do not consistently surpass existing SLMs. 
Further analysis suggests that the underperformance is largely attributed to the challenges posed by the numerous predefined relations.
To overcome this issue, we propose an LLM-based \underline{H}ierarchical \underline{C}lassification model for cross-document \underline{RE} (HCRE), 
which consists of two core components: 
1) an LLM for relation prediction 
and 
2) a \textit{hierarchical relation tree} derived from the predefined relation set. 
This tree enables the LLM to perform hierarchical classification, where the target relation is inferred level by level. 
Since the number of child nodes is much smaller than the size of the entire predefined relation set, the hierarchical relation tree significantly reduces the number of relation options that LLM needs to consider during inference. 
However, hierarchical classification introduces the risk of error propagation across levels. 
To mitigate this, we propose a \textit{prediction-then-verification} inference strategy that improves prediction reliability through multi-view verification at each level. 
Extensive experiments show that HCRE outperforms existing baselines, validating its effectiveness. 
We release our code at \url{https://github.com/XMUDeepLIT/HCRE}. 

\end{abstract}

\section{Introduction}
As a fundamental NLP task, relation extraction (RE) aims to extract structured relational triples 
from unstructured texts. 
In this aspect, early studies~(\citealp{zeng-etal-2014-cnn-re}; \citealp{cai-etal-2016-rnn-re}; \citealp{yao-etal-2019-docred}; \citealp{zhang-etal-2023-hypernetwork}; \citealp{zhang-etal-2024-multi-level}; \citealp{zhang-etal-2025-self}), mainly focus on identifying target relations 
between the head and tail entities
from a single sentence or document. 
However, numerous relational facts 
involve head and tail entities that do not co-occur in the same sentence or document. 
For instance, 
\citet{yao-etal-2021-codred} indicates that over half of Wikidata's relational facts span multiple documents.
Consequently, many researchers have shifted their attention to cross-document RE. 
Unlike conventional RE, cross-document RE requires the comprehensive analysis of multiple lengthy documents to predict relations between entities located in different documents.

\begin{figure}
    \centering
    \includegraphics[width=1\linewidth]{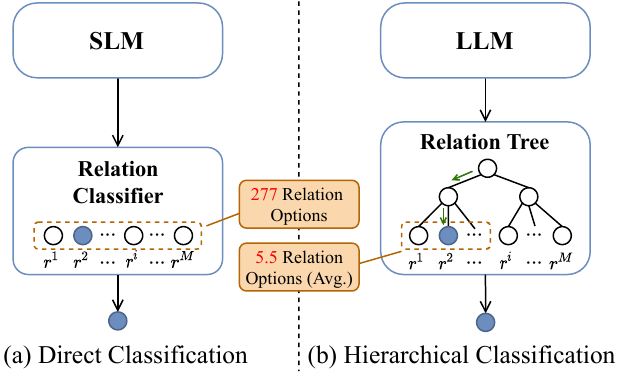}
    \caption{
        Comparison between the paradigm of existing cross-document RE model (Direct Classification)~and ours (Hierarchical Classification). 
        Green arrows indicate hierarchical classification based on the relation tree. 
    }
    \label{fig:ms}
\end{figure}

Existing approaches to cross-document RE can be roughly divided into two categories. 
The first category of studies~\cite{yao-etal-2021-codred, wang-etal-2022-ecrim, lu-etal-2023-mrcod, son-etal-2023-pilot, na-etal-2024-reic} primarily explore methods for extracting evidential context from lengthy documents, 
enabling the Small Language Models (SLMs) to effectively identify cross-document relations.
The second category of studies~\cite{yao-etal-2021-codred, wu-lgcr, yue-etal-2024-nepd} mainly focus on modeling long-range dependencies  
between head and tail entities that are connected through relevant entities within the documents. 
Despite their progress, these approaches remain confined to the paradigm of ``\textit{SLM + Classifier}''.
As illustrated in Figure~\ref{fig:ms}(a), this paradigm requires models to directly select the target relation from an extensive predefined relation set, which poses significant challenges for SLMs due to their limited language understanding capability, thereby hindering further performance improvement.

Given the strong capabilities demonstrated by Large Language Models (LLMs) in various NLP tasks~\cite{wang-etl-2023-instructuie, wadhwa-etal-2023-revisiting, li-etal-2023-codeie, zhou-etal-2024-grasping, shao-etal-2024-one2set-llm, tu-etal-2025-llm, lian2026sweagilesoftwareagentframework}, applying LLMs to cross-document RE appears to be a promising research direction. 
Driven by this goal, we conduct a preliminary study to explore the performance of LLMs in cross-document RE.
Our findings indicate that the performance of LLMs is limited and sometimes even lower than that of strong SLMs. 
Further analysis reveals that this suboptimal performance is largely attributable to the difficulty LLMs face when handling a large number of predefined relations in cross-document RE.
On the one hand, an excessive number of relations hinders the model's ability to distinguish between semantically similar relations. 
On the other hand, listing all predefined relations unnecessarily increases the context length, diverting the attention of LLMs away from critical details in documents.

Based on this insight, 
we propose an LLM-based \underline{H}ierarchical \underline{C}lassification model
for cross-document \underline{RE} (HCRE). 
It consists of two core components: 
1) an LLM used for relation prediction and 2) a \textit{hierarchical relation tree} that narrows down the relation options during prediction. 
We carefully design a pipeline to construct a hierarchical relation tree, where each leaf node corresponds to a predefined relation and intermediate nodes represent higher-level concepts of their respective children. 
Guided by this tree, 
our LLM performs \textit{hierarchical classification}, inferring a target relation in a top-down, level-by-level manner, as shown in Figure~\ref{fig:ms}(b). 
In this way, we present the LLM with only a small-scale set of tree nodes as relation options at each level,
thereby alleviating the challenge posed by a large relation set for the LLM.

Although hierarchical classification mitigates the issue of excessive predefined relations, it also introduces the issue of error propagation across levels.  
To alleviate this issue, we propose a level-wise \textit{prediction-then-verification} inference strategy. 
This strategy consists of two steps: 
1) \textit{Prediction step}. 
The LLM selects the best node and a suboptimal node from the given options based on the input instance. 
2) \textit{Verification step}. 
Through replacing selected nodes with their children, we further construct multiple verification option sets.
These sets are then fed back to the LLM to obtain additional predictions, which are used to verify the result from the prediction step. 
By iterating the above two steps, our strategy eliminates error-prone options and obtains a more accurate prediction at each level, effectively mitigating error propagation during hierarchical classification. 

To summarize, our contributions are threefold: 
{
\begin{itemize}
    \item To the best of our knowledge, our work is the first to explore the LLM-based hierarchical classification for cross-document RE. To investigate the performance of LLMs in this task, we identify a critical issue: the excessive number of relation options is the primary bottleneck limiting LLM performance in cross-document RE. This finding not only motivates our framework design but also provides a research direction for the community.
    \item To address the crucial issue, we propose a LLM-based \textit{hierarchical classification} model for cross-document RE, which reduces the number of relation options via a hierarchical relation tree, effectively improving the performance of LLMs in cross-document RE.
    \item To alleviate error propagation during hierarchical classification, we propose a level-wise \textit{prediction-then-verification} inference strategy, which improves prediction reliability through multi-view verification at each tree level. 
\end{itemize}
}

\section{Preliminary Study}

In this section, we conduct a preliminary study on the CodRED dataset to investigate the challenges of directly applying LLMs to cross-document RE, including evaluation metrics and model performance. 

\subsection{Challenges in Evaluation Metrics}

To analyze the reliability of commonly used metrics, we conduct a series of experiments on the CodRED benchmark~\cite{yao-etal-2021-codred} using several representative RE models, including End-to-End~\cite{yao-etal-2021-codred}, ECRIM~\cite{wang-etal-2022-ecrim}, and NEPD~\cite{yue-etal-2024-nepd}. 
To be specific, we investigate four evaluation metrics: 
1) \textbf{Maximum F1.} This metric denotes the maximum F1 score on the Precision-Recall curve. 
2) \textbf{P@K.} It measures model precision among the K relation triples with the highest confidence. 
3) \textbf{Micro F1.} 
This classic metric
provides an accurate measure for the trade-off between precision and recall across different relation classes. 
4) \textbf{Binary F1.} A variant of micro F1, this metric treats all positive relations as a single positive class and considers the NA (Not Available) label as the negative class, 
aiming to assess the ability of RE models in discriminating whether any relation exists between target entities.

\begin{table}[tbp]
\setlength{\tabcolsep}{0.5em}
\centering
\begin{tabular}{cccc}
\toprule
\textbf{Baselines} & \textbf{max F1} & \textbf{micro F1} & \textbf{binary F1} \\ \midrule
End-to-End         & 49.07               & 33.33             & 41.76              \\
ECRIM              & 61.64               & 39.25             & 42.98              \\
NEPD               & 63.53               & 25.77             & 32.00              \\ \bottomrule
\end{tabular}
\caption{Comparison among maximum F1, micro F1 and binary F1. }
\label{tab:pre-exp-metrics}
\end{table}

\begin{figure}[tbp]
    \centering
    \includegraphics[width=0.99\linewidth]{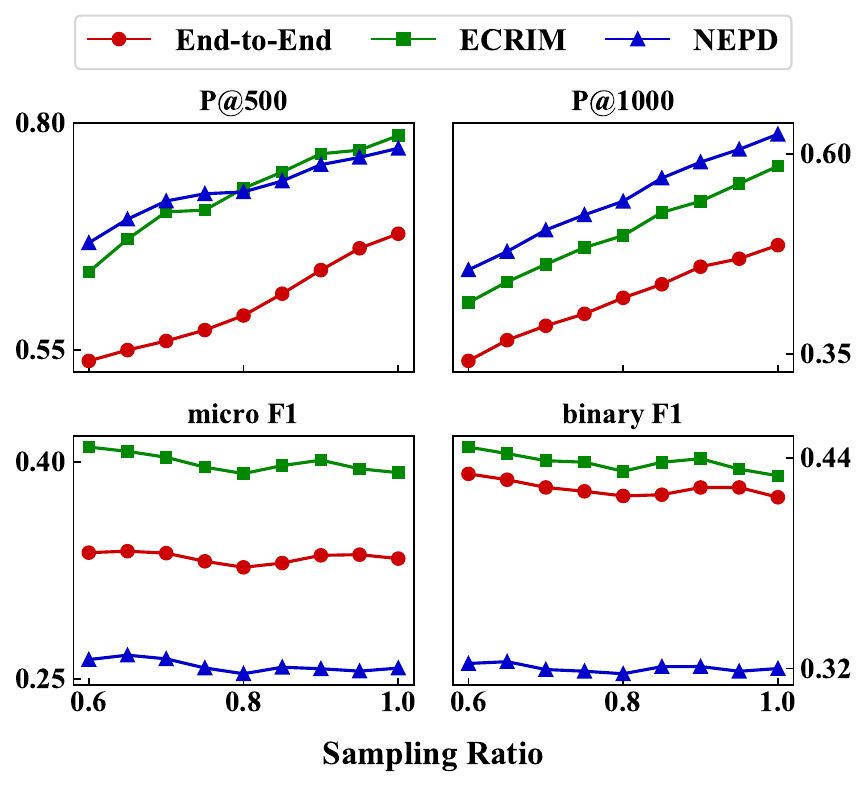}
    \caption{
        Comparison among P@K, micro F1, and binary F1 under varying dataset sizes. 
    }
    \label{fig:pre-exp-metrics}
\end{figure}

\begin{figure}[tbp]
    \centering
    \includegraphics[width=0.99\linewidth]{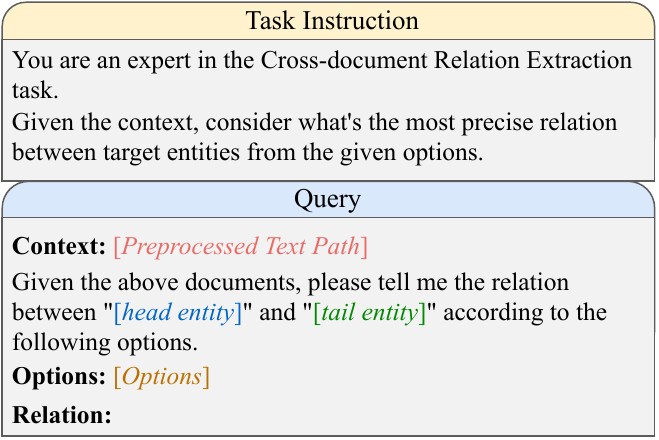}
    \caption{
        Our prompt template. 
        Each prompt contains the preprocessed text path, head and tail entities, and relation options. 
    }
    \label{fig:prompts}
\end{figure}

First, we review the calculation of \textbf{maximum~F1}, which requires adjusting the decision threshold to achieve the best balance between precision and recall. 
Since the optimal threshold is unavailable in real-world scenarios, maximum F1 fails to reflect the actual model ability and instead only reflects the ideal performance. 
Therefore, relying on maximum F1 may lead to an overestimation of the model's practical performance. 
From Table~\ref{tab:pre-exp-metrics}, we can observe that maximum F1 scores consistently exceed micro F1 and binary F1 across all models, 
further demonstrating its tendency to overestimate model performance.

Second, to analyze the effect of evaluation dataset scale on 
\textbf{P@K},
we compare the model performance on subsets of different sizes sampled from the development set of CodRED. 
From Figure~\ref{fig:pre-exp-metrics}, we observe that:
1) P@K increases with larger dataset size, indicating that P@K is sensitive to dataset scale; 
2) P@K (e.g., P@500) may yield inconsistent model performance rankings under varying evaluation dataset scales. 
These results suggest that P@K is not sufficiently reliable for real-world scenarios, where the number of test instances is not fixed.
In contrast, micro F1 and binary F1 remain consistently stable across all dataset scales, demonstrating greater reliability for evaluation.

Third, maximum F1 and P@K require assigning scores to all predefined relations simultaneously, whereas LLMs are trained to generate relation labels token by token. This mismatch makes such metrics unsuitable for LLMs.

Based on the above analysis, we adopt \textbf{micro~F1} and \textbf{binary F1} as our primary evaluation metrics in subsequent experiments for the following reasons: 
1) Both metrics avoid the flaws discussed earlier. 
2) Given the large proportion of negative instances in cross-document RE  
\cite{yao-etal-2021-codred, yue-etal-2024-nepd}, it is crucial for RE models to distinguish positive instances from negative ones, which motivates our use of binary F1.

\subsection{Challenges in Model Performance}

\begin{figure}[tbp]
    \centering
    \includegraphics[width=0.99\linewidth]{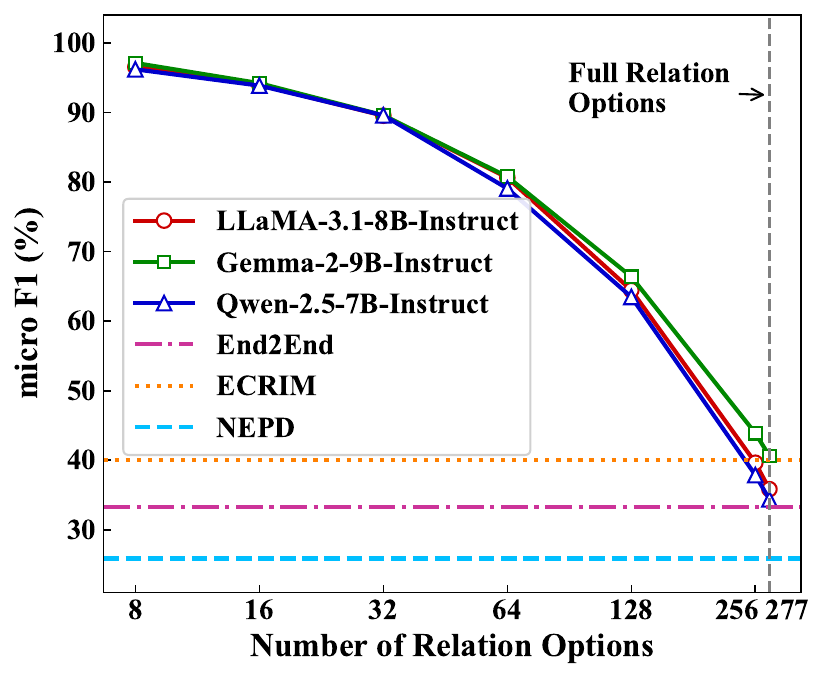}
    \caption{
        Effect of the number of relation options on LLMs. 
        Note that CodRED involves 277 predefined relations. 
        The horizontal dashed lines denote SLM performance serving as baselines. 
    }
    \label{fig:option nums}
\end{figure}

Here, we explore the performance of applying an LLM directly to cross-document RE. 
To this end, we choose 
LLaMA-3.1-8B-Instruct \cite{grattafiori-2024-llama3}, 
Gemma-2-9B-Instruct \cite{gemmateam2024gemma2improvingopen}, 
and Qwen-2.5-7B-Instruct \cite{qwen2025qwen25technicalreport}
as the representative LLMs and compare them with the previous SLM-based baselines. 
All models are fine-tuned on the training set of CodRED dataset and micro F1 scores on the development set are reported.
As shown in Figure~\ref{fig:prompts}, we employ a prompt template to transform each input instance into a text prompt, 
thereby guiding LLMs to directly generate the target relation. 

As depicted by the gray dashed line in Figure~\ref{fig:option nums}, 
the performance of directly fine-tuned LLMs is unsatisfactory, sometimes even underperforming strong SLM-based baselines. 
We argue that the suboptimal performance of LLMs is attributed to the massive predefined relations in cross-document RE, 
which makes it difficult for LLMs to distinguish similar relations and leads to an overlong input prompt, 
thereby interfering with model prediction. 

To verify our hypothesis, 
we further assess their performance with varying numbers of relation options. 
Specifically, during both training and inference, we randomly remove a certain number of relation options from the original prompt for each instance.
As illustrated in Figure~\ref{fig:option nums}, reducing the number of relation options leads to a significant performance improvement in LLMs. 
This motivates us to explore methods that reduce the number of relation options to enhance model predictions.

\section{Our Model}

\begin{figure*}[htbp]
    \centering
    \includegraphics[width=1\linewidth]{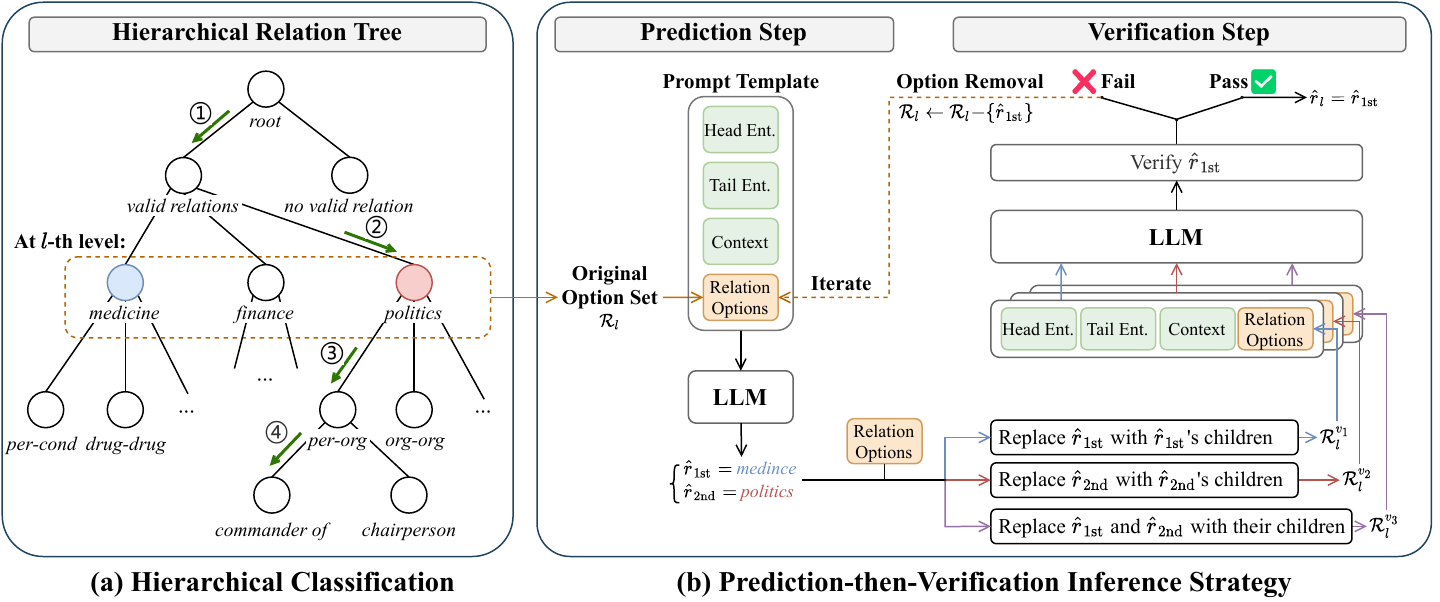}
    \caption{
        Overview of HCRE. 
        (a) Guided by the hierarchical relation tree, the LLM $\mathcal M_1$ performs \textit{hierarchical classification} in a top-down manner (See green arrows). 
        (b) At each level, we refine model prediction by \textit{prediction-then-verification} inference strategy, which is composed of the \textit{prediction step} and the \textit{verification step}. 
    }
    \label{fig:ptv}
\end{figure*}

Based on the findings from our preliminary study, we propose HCRE, an LLM-based hierarchical classification model for cross-document RE. 
In this section, we provide a comprehensive description of our hierarchical classification model, which consists of an LLM $\mathcal{M}_1$ and a hierarchical relation tree. 
Concretely, we first introduce a pipeline to construct a \textit{hierarchical relation tree} in Section~\ref{sec:lt}.
Based on this tree, we elaborate on the LLM-based \textit{hierarchical classification} process in Section~\ref{sec:hc}.
Finally, we detail the model inference and training strategies in Sections~\ref{sec:ptv}~and~\ref{sec:train}, respectively.

\subsection{Hierarchical Relation Tree}\label{sec:lt}

To address the performance degradation in LLM-based cross-document RE caused by numerous predefined relations, we construct a \textit{hierarchical relation tree} to effectively organize these relations.
To achieve this, we design a pipeline that employs an advanced LLM $\mathcal{M}_2$ (such as GPT-4o) to construct the tree based on the semantics of predefined relations.
In this pipeline, we initialize the hierarchical relation tree with a root node encompassing all predefined relations, and then recursively partition upper-level nodes to obtain lower-level nodes, constructing the hierarchy level by level.

To enhance the distinctiveness of nodes at each level, we prompt the LLM $\mathcal M_2$ to generate a textual partitioning criterion $C_l$ for each level $l$.
At the $l$-th level, we prompt $\mathcal M_2$ with $C_l$ to partition the relations contained in each current-level node, thereby producing child nodes at the $(l\text{+}1)$-th level. 
Then, $\mathcal M_2$ generates a node name for each child node, reflecting the common aspects of relations it contains under the criterion $C_l$. 
For example, using the criterion ``\textit{domain}'' enables us to split each node according to the domains of its corresponding relations, forming child nodes such as ``\textit{politics}'' and ``\textit{finance}''. 
This partitioning process continues until a maximum tree depth $L$ is reached. 

In addition, to alleviate the label imbalance in cross-document RE, we further impose a constraint on the second level of the tree, limiting it to two nodes: ``\textit{valid relations}'', which includes all positive relations, and ``\textit{no valid relation}'', which contains only NA.
This design explicitly separates positive relations from NA, enabling $\mathcal M_1$ to better distinguish between these two kinds of instances and reducing the model's bias towards predicting NA due to its overabundance in the training data.

Following the above pipeline, we obtain a hierarchical relation tree, where leaf nodes correspond to predefined relations and intermediate nodes represent higher-level concepts of their respective child nodes. 
More details regarding the tree are provided in \textbf{Appendix~\ref{ap:lt}}. 

\subsection{LLM-based Hierarchical Classification}\label{sec:hc}

After obtaining the hierarchical relation tree, for each instance $x$ (context, head entity, and tail entity), we guide the LLM $\mathcal M_1$ to infer its target relation at a leaf node by performing a top‑down, level‑by‑level \textit{hierarchical classification}, as shown in Figure~\ref{fig:ptv}(a).
At each level, $\mathcal M_1$ only needs to consider a small set of relation options, effectively avoiding suboptimal performance caused by an excessive number of relation options.

Since $\mathcal M_1$ performs similar computations at each level of the tree, we illustrate this procedure in detail using the $l$-th level as an example.
Here, $\hat r_{l-1}$ refers to the node chosen by $\mathcal M_1$ at the $(l\text{-}1)\text{-th}$ level, and $\mathcal R_l$ denotes the set of its child nodes at the $l$-th level.
Next, we use $\mathcal R_l$ as the set of relation options to instantiate the prompt template in Figure~\ref{fig:prompts}, thereby prompting $\mathcal M_1$ to select the optimal node $\hat r_l$ from $\mathcal R_l$ for the next level. 
This process is repeated until a leaf node corresponding to the target relation of instance $x$ is reached.
Crucially, the number of child nodes $|\mathcal R_l|$ for any parent is much smaller than the total number of predefined relations, significantly reducing the number of options $\mathcal M_1$ must consider during inference.

\subsection{Prediction-then-Verification Inference Strategy}\label{sec:ptv}

Although hierarchical classification effectively reduces the number of relation options considered by $\mathcal M_1$ per level, errors occurring at earlier levels may propagate and affect predictions at subsequent levels. 
To address the issue of error propagation, we propose a \textit{prediction-then-verification} inference strategy to refine model prediction at each level through multi-view verification. 
As illustrated in Figure~\ref{fig:ptv}(b), our strategy enhances the reliability of model prediction at each level by iterating the following two steps:

\textbf{Prediction Step.} 
At the $l$-th level of the hierarchical relation tree, we first adopt the children of the node $\hat r_{l-1}$ predicted at the $(l\text{-}1)$-th level as the relation option set $\mathcal R_l$. 
Then, $\mathcal M_1$ selects the best and suboptimal nodes from $\mathcal R_l$, denoted as $\hat r_\text{1st}$ and $\hat r_\text{2nd}$, respectively.
Intuitively, $\mathcal M_1$ tends to confuse $\hat r_\text{1st}$ and $\hat r_\text{2nd}$, which may lead to an incorrect prediction.

\textbf{Verification Step.} 
To mitigate the confusion between $\hat r_\text{1st}$ and $\hat r_\text{2nd}$, we validate the reliability of $\hat r_\text{1st}$ using multiple \textit{verification option sets}. 
By treating child nodes as a finer-grained alternative view of their parent node, we replace each node with its children to construct these sets.
Specifically, we replace $\hat r_\text{1st}$ and $\hat r_\text{2nd}$ with their respective child nodes in the original relation option set $\mathcal R_l$, forming two verification option sets $\mathcal R_l^{v_1}$ and $\mathcal R_l^{v_2}$. 
Meanwhile, the third verification option set $\mathcal R_l^{v_3}$ is obtained by simultaneously replacing both best and suboptimal nodes with their child nodes in $\mathcal R_l$. 

In this way, we obtain three alternative views of $\mathcal R_l$, which allow us to verify whether $\mathcal M_1$ consistently predicts nodes that are semantically aligned with the best node $\hat r_\text{1st}$ (i.e., exactly matches $\hat r_\text{1st}$ or one of its children). 
Concretely, $\mathcal M_1$ is prompted to select the best node from each verification option set to serve as an \textit{auxiliary verification node}.
If more than half of the auxiliary verification nodes are semantically aligned with $\hat r_\text{1st}$, $\hat r_\text{1st}$ is considered as a reliable and final prediction $\hat r_l$ at the current level, and then we proceed to the next level of the hierarchical relation tree. 
Otherwise, $\hat r_\text{1st}$ is deemed incorrect and removed from $\mathcal R_l$, after which we repeat the above two steps. 

It should be noted that $\mathcal R_l^{v_1}$, $\mathcal R_l^{v_2}$ and $\mathcal R_l^{v_3}$ are essentially three equivalent views of $\mathcal R_l$ at a finer granularity.
During the verification step, by incorporating next-level nodes, the verification option sets provide finer-grained semantic information compared to the original relation option set $\mathcal R_l$. 
This finer granularity enables $\mathcal M_1$ to effectively discern subtle differences between $\hat r_\text{1st}$ and $\hat r_\text{2nd}$, ultimately resulting in a more reliable prediction.
Further elaboration on the prediction-then-verification inference strategy is provided in \textbf{Appendix~\ref{ap:ptv-more}}.

\subsection{Model Training}\label{sec:train}

To effectively train our LLM for hierarchical classification in cross-document RE, we reconstruct the training samples accordingly.

For each training sample $(x, \mathcal R, r)$, we begin by identifying a path $r_{0}, r_{1}, \cdots, r_{L-1}$ from the root node to the leaf node $r_{L-1}\text{ = }r$ within the hierarchical relation tree.
Here, $x$, $\mathcal R$, $r$, and $L$ represent the input instance, predefined relation set, target relation, and tree depth, respectively. 
Utilizing this path, we extend $(x, \mathcal R, r)$ to $L\text{-}1$ level-wise training samples $\{(x, \mathcal R_l, r_{l})\}_{l=1}^{L-1}$, 
where $\mathcal R_l$ denotes the relation option set containing $r_{l}$ and its siblings.
This process is repeated for all training samples to form a new dataset $\mathcal D_1$.

On top of that, to simulate the verification step, we construct another dataset $\mathcal D_2$.
For each training sample $(x, \mathcal R_l, r_l)$ in $\mathcal D_1$, three new training samples are derived: $\{(x, \mathcal R^{v_1}_l, r_{l+1}),$ $ (x, \mathcal R^{v_2}_l, r_l), $ $(x, \mathcal R^{v_3}_l, r_{l+1})\}$. 
During the construction process, the ground-truth node is treated as the best node $r_\text{1st}$, and a randomly selected sibling node serves as the suboptimal node $r_\text{2nd}$. 
These two nodes are then replaced with their respective child nodes to form the verification option sets $\mathcal R^{v_1}_l$, $\mathcal R^{v_2}_l$, and $\mathcal R^{v_3}_l$. 
Finally, our LLM $\mathcal M_1$ is fine-tuned on the combined dataset $\mathcal D_1 \cup \mathcal D_2$.

\section{Experiments}

\subsection{Experiment Setup}\label{sec:setup}

\textbf{Datasets. }
We conduct our experiments on CodRED~\cite{yao-etal-2021-codred}, a widely used benchmark for cross-document RE. 
CodRED provides two settings: the closed setting offers gold text paths in advance, whereas the open setting requires retrieving relevant text paths for relation prediction.
Detailed statistics of CodRED are provided in Table~\ref{tab:stats-codred}. 

\noindent\textbf{Baselines. }
We primarily compare our model with two kinds of baselines: 
1) \textit{Cross-document RE Baselines}, including \textbf{End-to-End}~\cite{yao-etal-2021-codred}, \textbf{ECRIM}~\cite{wang-etal-2022-ecrim}, \textbf{MR.COD}~\cite{lu-etal-2023-mrcod}, \textbf{KXDocRE}~\cite{jain-etal-2024-kxdocre}, \textbf{REIC}~\cite{na-etal-2024-reic}, and \textbf{NEPD}~\cite{yue-etal-2024-nepd}; 
and 2) \textit{LLM-based Hierarchical Text Classification (HTC) Baselines}, including \textbf{Rs-ICL}~\cite{chen-etal-2024-rsicl}, \textbf{DFS-L.}~\cite{yu2022constrained} and \textbf{BFS-L.}~\cite{huang2022exploring, jain-etal-2024-higen}. 
Moreover, we include a directly fine-tuned baseline, denoted as \textbf{Vanilla}.
The details of these baselines are provided in \textbf{Appendix~\ref{ap:baseline}}. 

In addition, the implementation details of our model are provided in \textbf{Appendix~\ref{ap:impl}}. 


\begin{table}[tbp]
\setlength{\tabcolsep}{0.5em}
\centering
\begin{tabular}{lccc}
\toprule
\multirow{2}{*}{}  & \multicolumn{2}{c}{\textbf{Closed}} & \textbf{Open}   \\
                   & \textbf{Train}        & \textbf{Dev}          & \textbf{Dev}    \\ \midrule
\#Text Path (Pos.) & 8,263        & 2,558        & 4,558  \\
\#Text Path (NA)   & 120,925      & 38,182       & 15,072 \\
\#Tokens / Doc     & 4,938.6      & 5,031.6      & 62,863 \\ \bottomrule
\end{tabular}
\caption{Statistics of CodRED. (Pos.: Positive; NA: Not Available.)}
\label{tab:stats-codred}
\end{table}

\begin{table*}[!ht]
\setlength{\tabcolsep}{0.42em}
\centering
\begin{tabular}{clcccccc}
\toprule
\multirow{2}{*}{\textbf{Backbone}} & \multicolumn{1}{c}{\multirow{2}{*}{\textbf{Model}}} & & \multicolumn{2}{c}{\textbf{Closed}} & & \multicolumn{2}{c}{\textbf{Open}} \\
& & & \textbf{micro F1} & \textbf{binary F1} & & \textbf{micro F1} & \textbf{binary F1} \\ \midrule
\multicolumn{8}{c}{\textbf{\textit{Cross-Document RE Baselines}}} \\ \midrule
\multicolumn{1}{c}{\multirow{5}{*}{BERT-base}}
& \multicolumn{1}{l}{End-to-End~\cite{yao-etal-2021-codred}} &  & 33.33 & 41.76 & & 22.23 & 28.85 \\
& \multicolumn{1}{l}{ECRIM~\cite{wang-etal-2022-ecrim}} & & 39.25 & 42.98 & & 19.81 & 22.06 \\
& \multicolumn{1}{l}{KXDocRE~\cite{jain-etal-2024-kxdocre}} & & 37.53 & 38.25 & & 18.80 & 19.39 \\
& \multicolumn{1}{l}{REIC~\cite{na-etal-2024-reic}} & & 38.75 & 46.25 & & 19.50 & 23.77 \\
& \multicolumn{1}{l}{NEPD~\cite{yue-etal-2024-nepd}} & & 25.77 & 32.00 & & 29.16 & 36.36 \\
&&&&&&&\\[-1.8ex]
\hdashline[5pt/3pt]
&&&&&&& \\[-1.6ex]
\multicolumn{1}{c}{\multirow{5}{*}{RoBERTa-large}}
& \multicolumn{1}{l}{End-to-End~\cite{yao-etal-2021-codred}} & & 41.45 & 47.99 & & 25.35 & 29.60 \\
& \multicolumn{1}{l}{ECRIM~\cite{wang-etal-2022-ecrim}} & & 42.54 & 49.47 & & 23.39 & 27.60 \\
& \multicolumn{1}{l}{KXDocRE~\cite{jain-etal-2024-kxdocre}} & & 42.55 & 45.36 & & 21.74 & 23.23 \\
& \multicolumn{1}{l}{REIC~\cite{na-etal-2024-reic}} & & 40.17 & 48.74 & & 22.49 & 27.97 \\
& \multicolumn{1}{l}{NEPD~\cite{yue-etal-2024-nepd}} & & 42.96 & 52.67 & & 30.12 & 37.04 \\
\midrule
\multicolumn{8}{c}{\textbf{\textit{LLM-based HTC Baselines}}} \\ \midrule
\multicolumn{1}{c}{GPT-4o-mini} 
& \multicolumn{1}{l}{Rs-ICL~\cite{chen-etal-2024-rsicl}} & & 11.22 & 39.26 & & 10.55 & 39.86 \\ 
&&&&&&&\\[-1.8ex]
\hdashline[5pt/3pt]
&&&&&&& \\[-1.6ex]
\multicolumn{1}{c}{\multirow{2}{*}{LLaMA-3.1-8B}}
& \multicolumn{1}{l}{DFS-L.~\cite{yu2022constrained}} & & 36.83 & 42.74 & & 14.51 & 18.09 \\
& \multicolumn{1}{l}{BFS-L.~\cite{jain-etal-2024-higen}} & & 35.55 & 41.95 & & 13.46 & 16.46 \\ \midrule
\multicolumn{8}{c}{\textbf{\textit{Ours}}} \\ \midrule
\multicolumn{1}{c}{\multirow{2}{*}{LLaMA-3.1-8B}} 
& \multicolumn{1}{l}{Vanilla} & & 38.14 & 41.43 & & 15.19 & 17.00 \\
& \multicolumn{1}{l}{HCRE} & & \textbf{45.35\textsubscript{3}\ddag} & \textbf{58.19\textsubscript{3}\ddag} & & \textbf{34.91\textsubscript{2}\ddag} & \textbf{49.33\textsubscript{2}\ddag} \\ \bottomrule
\end{tabular}
\caption{
    Experimental results on CodRED under both closed and open settings. 
    The subscript denotes the corresponding standard deviation (e.g., 45.35\textsubscript{3} represents 45.35 $\pm$ 0.3). 
    \ddag\ indicates significance at $p < 0.01$ over the second-best baseline NEPD, based on 1,000 bootstrap tests \cite{tibshirani1993bootstrap}.
}
\label{tab:main-results}
\end{table*}

\begin{table}[!ht]
\setlength{\tabcolsep}{0.75em}
\centering
\begin{tabular}{@{}lcc}
\toprule
\multicolumn{1}{c}{\textbf{Model}}   & \textbf{micro F1} & \textbf{binary F1} \\ \midrule
\multicolumn{1}{l}{Ours}                                 & 45.35             & 58.19              \\ \midrule
\multicolumn{1}{l}{\hspace{1em} \textit{w/o} multi-view} & 39.37             & 49.63              \\
\multicolumn{1}{l}{\hspace{1em} \textit{w/o} PtV}        & 37.66             & 47.28              \\
\multicolumn{1}{l}{\hspace{1em} \textit{w/o} LTC}        & 43.18             & 56.60              \\
\multicolumn{1}{l}{\hspace{1em} \textit{w/o} LTC, PtV}   & 32.55             & 45.44              \\
\multicolumn{1}{l}{\hspace{1em} \textit{w/o} HRT}        & 38.14             & 41.43              \\ \bottomrule
\end{tabular}
\caption{
    Ablation study of our model on CodRED under the closed setting. 
    Note that LTC, PtV, and HRT denote level-wise tree construction, prediction-then-verification, and hierarchical relation tree, respectively. 
}
\label{tab:ablation}
\end{table}


\subsection{Main Results}

The experimental results on both closed and open settings of CodRED are shown in Table~\ref{tab:main-results}. 
Based on these results, we draw several key conclusions: 

First, our model consistently outperforms all SLM-based baselines under both settings, demonstrating the potential of LLMs for cross-document RE. 
In particular, compared to the strongest SLM-based baseline, RoBERTa + NEPD, our model achieves gains of 2.39 and 5.52 points in micro F1 and binary F1 under the closed setting.

Second, in comparison with LLM-based HTC baselines, our model consistently achieves better performance. 
This demonstrates the effectiveness of our prediction-then-verification inference strategy for hierarchical classification by mitigating error propagation.

Third, while the Vanilla baseline does not outperform all the baselines, our model improves performance based on it and surpasses all the baselines. 
This suggests that reducing the number of relation options can effectively enhance the performance of LLMs in cross-document RE.

\subsection{Ablation Study}

In Table~\ref{tab:ablation}, we investigate the effects of each component in our model to verify their effectiveness. 
Specifically, we compare our model with the following variants:

(1) \textit{w/o multi-view}. 
In this variant, instead of adopting three verification option sets $\{\mathcal R^1, \mathcal R^2, \mathcal R^3\}$, 
we only adopt the first option set $\mathcal R^1$ for verification. 
which leads to a notable performance drop. 
This result highlights the importance of incorporating fine-grained information from multiple views, which enables the LLM $\mathcal M_1$ to verify the reliability of predictions more effectively. 

(2) \textit{w/o PtV}. 
The removal of the prediction-then-verification (PtV) inference strategy from HCRE results in a substantial performance drop.
This underscores the critical role of the verification step in hierarchical classification, as it improves overall performance by mitigating errors at each level. 

(3) \textit{w/o LTC}. 
Here, we simply prompt $\mathcal M_2$ to directly generate the hierarchical relation tree based on the predefined relations, rather than constructing the relation tree level by level.
In this variant, we observe a slight performance drop.
This is potentially because the hierarchical relation tree constructed by our level-wise pipeline comprises more distinguishable nodes, allowing LLMs to identify target relations more accurately.  
More details about this variant are in \textbf{Appendix~\ref{ap:vanilla-lt-build}}. 

(4) \textit{w/o LTC, PtV}. 
In this variant, we further remove the PtV strategy from the \textit{w/o LTC} variant and observe a more significant performance degradation. 
We attribute this to the stronger semantic relevance between parent and child nodes in our tree, which enables child nodes to better represent their parent node during the verification step.

(5) \textit{w/o HRT}. 
Different from the above variants, this variant requires $\mathcal M_1$ to select target relations from the entire predefined relation set.
A slight performance decline is observed compared to the \textit{w/o PtV} variant, suggesting $\mathcal M_1$ makes better relation prediction, benefiting from the hierarchical relation tree that reduces the number of relation options considered during inference. 

\subsection{Analysis of Error Propagation}

\begin{figure}[tbp]
    \centering
    \includegraphics[width=0.99\linewidth]{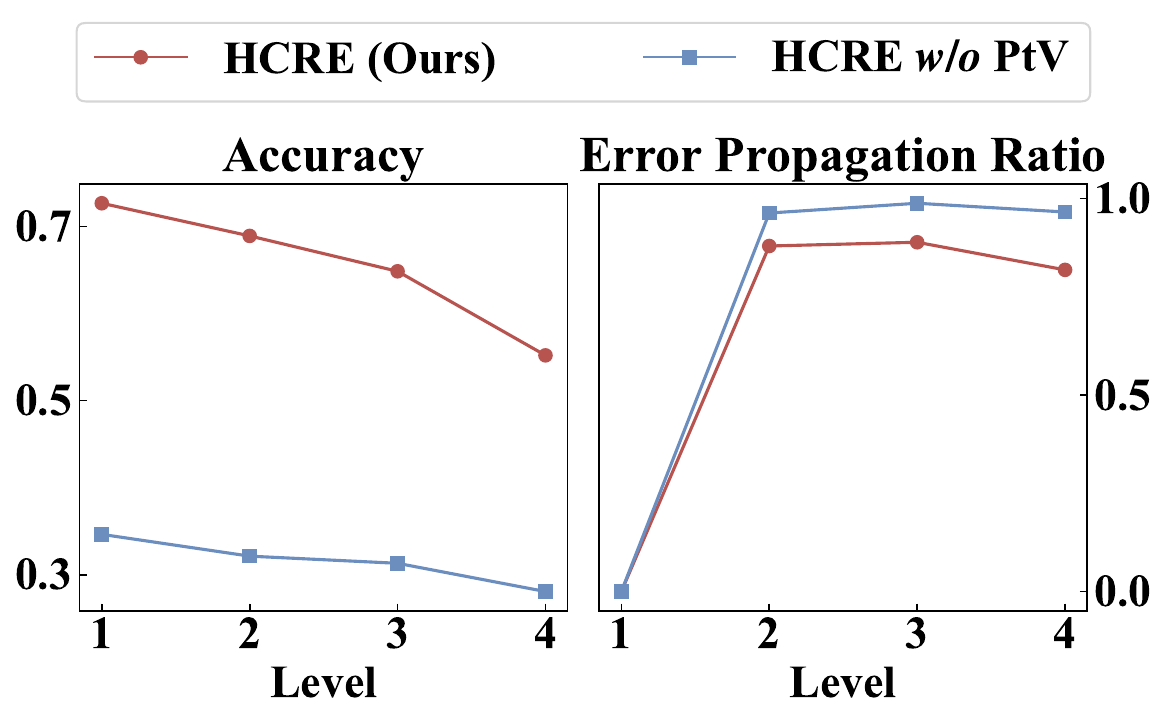}
    \caption{
        The accuracy and error propagation ratio of our model, with and without the PtV strategy. 
    }
    \label{fig:errp}
\end{figure}

To verify that the PtV strategy effectively mitigates error propagation during hierarchical classification, we evaluate the accuracy of our model at each level. 
Additionally, we focus on the misclassification instances caused by incorrect predictions at previous levels and define the proportion of such instances among all misclassifications as \textit{error propagation ratio}.
As illustrated in Figure~\ref{fig:errp}, the PtV strategy not only consistently improves model performance but also mitigates error propagation at all levels. 

\subsection{Effect of Tree Depth \texorpdfstring{$L$}{L}}\label{sec:tree-dep}

\begin{figure}[tbp]
    \centering
    \includegraphics[width=0.99\linewidth]{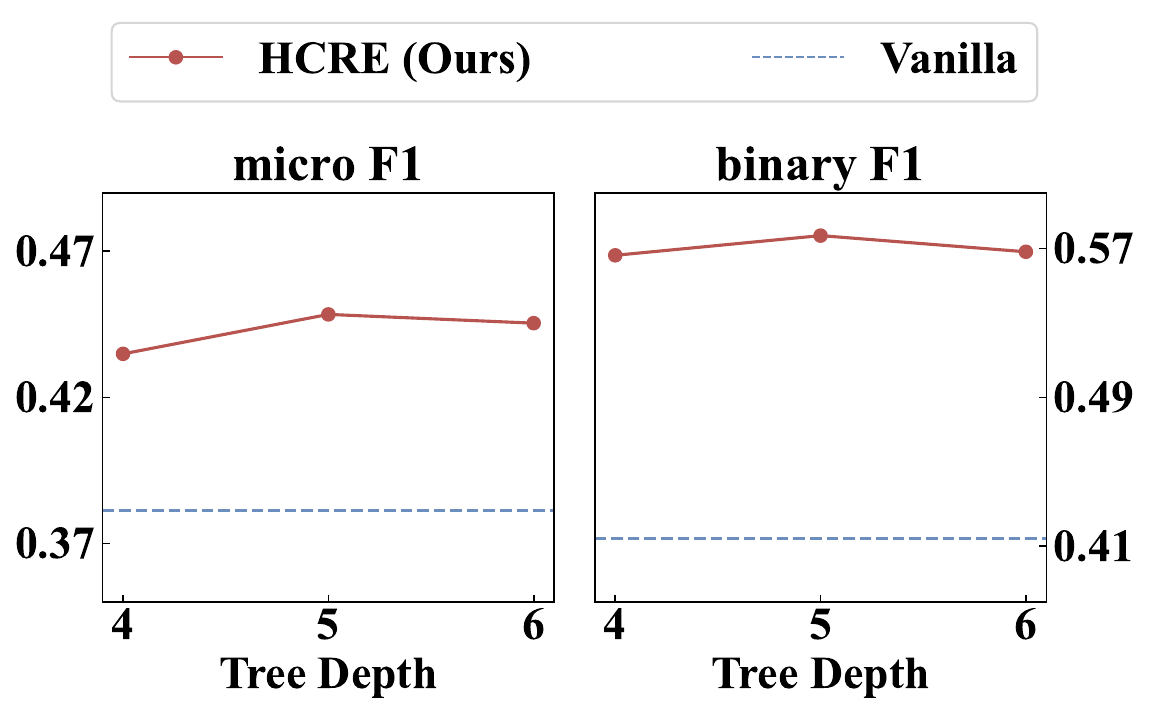}
    \caption{
        Model performance on the CodRED development set using trees with varying tree depth~$L$. 
    }
    \label{fig:tdepth}
\end{figure}

To investigate the impact of predefined tree depth~$L$, we evalute our model using relation trees with $L \in \{\text{4}, \text{5}, \text{6}\}$.
Figure~\ref{fig:tdepth} shows that HCRE is not sensitive to the predefined tree depth, exhibiting strong robustness. 
Since our model achieves the best performance at $L\text{ = }$5, we adopt 5 as the relation tree depth in all experiments.

We additionally present 
analysis of error propagation, 
bag-level evaluation, 
experiments on conventional evaluation metrics, 
studies using different backbone architectures, 
cross-dataset generalization results of HCRE, 
and 
more ablation experiments about PtV 
in 
\textbf{Appendices}~\textbf{\ref{ap:err-trans}}, 
\textbf{\ref{ap:bag}}, 
\textbf{\ref{ap:conv-metric}}, 
\textbf{\ref{ap:backbone}}, 
\textbf{\ref{ap:docred}}, 
and~\textbf{\ref{ap:ptv-more}}, 
respectively.

\section{Related Works}

\subsection{Cross-document Relation Extraction. }
Cross-document RE aims to identify the relations between entities appearing in different documents, a task first introduced by \citet{yao-etal-2021-codred}.
Subsequent studies mainly advance this task along two directions: evidence retrieval and long-range relation representation.
For evidence retrieval, prior work aims to reduce irrelevant context by selecting informative sentences or paths, such as document filtering based on entity co-occurrence \citep{wang-etal-2022-ecrim}, graph-based evidence path mining \citep{lu-etal-2023-mrcod}, text path expansion via bridge entities \citep{son-etal-2023-pilot}, and reinforcement learning-based sentence selection \citep{na-etal-2024-reic}.
Another line of research focuses on modeling long-range dependencies.
Representative approaches include cross-path relation attention \citep{wang-etal-2022-ecrim}, local-to-global causal reasoning over text paths \citep{wu-lgcr}, unified entity graph construction \citep{yue-etal-2024-nepd}, and domain knowledge injection \citep{jain-etal-2024-kxdocre}.
Despite their progress, their models still follow the ``\textit{SLM + Classifier}'' paradigm, constrained by the limited language understanding capabilities of SLMs. 
In contrast, our work explores LLM-based hierarchical classification for cross-document RE and avoids this flaw.

\subsection{LLM-based Hierarchical Text Classification. }
In recent years, LLMs have shown strong capabilities in 
HTC, which aims to predict labels organized in a hierarchy. 
Some works~\cite{wang-etal-2023-tbhtc, paletto-etal-2024-lazshtc, zhang2025teleclass} adopt LLMs as data annotators, highlighting the application of LLMs to generate high-quality data and enrich label taxonomies for HTC. 
Other works explore utilizing LLMs to enhance HTC at inference time, 
training models to perform hierarchical classification either by converting label hierarchies into sequences using depth-first~\cite{yu2022constrained} or breadth-first orders~\cite{huang2022exploring, jain-etal-2024-higen} or by classifying level by level~\cite{jain-etal-2024-higen, chen-etal-2024-rsicl, tabatabaei-etal-2025-large}.
Despite the effectiveness of these methods, they often neglect error propagation during hierarchical classification. 
In contrast, our model mitigates this issue through multi-view verification at each level. 

\section{Conclusion and Future Work}

In this paper, we have proposed an LLM-based hierarchical classification model for cross-document RE. 
Specifically, we utilize the predefined relations to construct a hierarchical relation tree, which guides our LLM to infer target relations level by level. 
Furthermore, we propose a prediction-then-verification inference strategy to refine model predictions at each level, thereby mitigating error propagation. 
Empirical results on the commonly used benchmark CodRED show the superiority of HCRE over existing baselines. 
In the future, we plan to investigate the generalization ability of our model on more information extraction tasks.

\section*{Limitations}

Our study has two main limitations that warrant further investigation. 
First, due to the limitation of input context length, our LLM can only handle a single text path during each inference, which prevents the model from leveraging cross-path dependency information that could potentially enhance its performance. 
Second, during the training of our model, we randomly sample a node as an alternative to the suboptimal node, which may not be the optimal strategy.

\section*{Acknowledgments}
The project was supported by National Natural Science Foundation of China (No. 62276219), Natural Science Foundation of Fujian Province
of China (No. 2024J011001), and the Public Technology Service Platform Project of Xiamen (No.3502Z20231043). We also thank the reviewers
for their insightful comments.

\bibliography{custom}

\clearpage  

\appendix

\section{Details of Hierarchical Relation Tree}\label{ap:lt}

\begin{figure*}[htbp]
    \centering
    \includegraphics[width=1\linewidth]{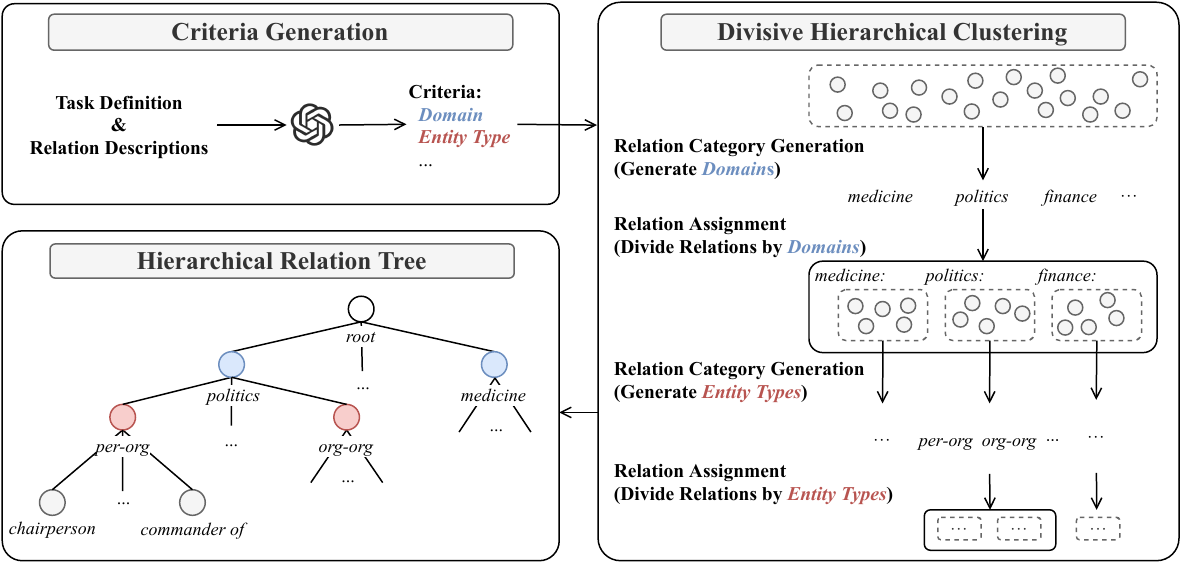}
    \caption{Illustration of the level-wise pipeline for hierarchical relation tree construction. }
    \label{fig:lt-new}
\end{figure*}

\subsection{Prompt Templates for Tree Construction}\label{ap:lt-prompt}

As illustrated in Figure~\ref{fig:lt-new}, our tree construction pipeline begins by deriving a set of partitioning criteria to progressively partition the relations. 
Using these criteria, we then recursively construct child nodes and populate them with relevant relations, constructing the hierarchy level by level. 

Specifically, we first adopt the following prompt template to generate partitioning criteria. 

\begin{tcolorbox}[
colback=gray!10, colframe=gray!90, 
title=Prompt Template for Criterion Generation, 
breakable]
\textbf{Background: }\\
You are an expert in cross-document relation extraction, which aims to identify predefined relations between entities that appear in different documents. \\
\textbf{Task: }\\
Your task is to analyze and provide several distinct classification criteria which is beneficial to group predefined relations based on: \\
  - Homogeneity: Relations in the same category should be highly similar.\\
  - Heterogeneity: Different categories should be as distinct as possible.\\
\textbf{Requirements: }\\
1. Provide 10-12 distinct classification criteria with concise names (1-2 words). \\
2. Choose the top 2 criteria that might yield the most effective classification results.\\
3. Return the result in a valid JSON format as shown below. \\
\textbf{JSON Output Format: }\\
\verb+```+json\\
\{\\
  ``classification criteria'': \{\\
    ``criterion name 1'': \{\\
      ``explanation'': ``explanation of criterion 1'', \\
      ``possible category names'': [ ``category name 1'', ``category name 2'' ]\\
    \}, \\
    ``criterion name 2'': \{\\
      ``explanation'': ``explanation of criterion 1'', \\
      ``possible category names'': [ ``category name 1'', ``category name 2'' ]\\
    \}, \\
    ...\\
  \}, \\
  ``top2 criteria'': [``criterion name i'', ``criterion name j'']\\
\}\\
\verb+```+\\
\textbf{Output: }
\end{tcolorbox}

Then, we employ the prompt template below to construct child nodes by dividing a parent node into multiple nodes. 
Here, \textcolor{brown}{[CRITERION\_NAME]} refers to a textual criterion, \textcolor{brown}{[CRITERION\_EXPLANATION]} denotes its corresponding description, and \textcolor{brown}{[RELATION\_WITH\_DESC]} represents the relations contained in the parent node along with their descriptions.

\begin{tcolorbox}[
colback=gray!10, colframe=gray!90, 
title=Prompt Template for Node Name Generation, 
breakable]
\textbf{Task: }\\
Analyze the given relation types and their descriptions, and categorize these relations into 10-12 clusters based on their \textcolor{brown}{[CRITERION\_NAME]}s, where ``\textcolor{brown}{[CRITERION\_NAME]}'' is defined as: ``\textcolor{brown}{[CRITERION\_EXPLANATION]}''. \\
\textbf{Requirements: } \\
1. Each \textcolor{brown}{[CRITERION\_NAME]} should have a \textbf{CONCISE, CLEAR and STRUCTURED} name (1-2 words) reflecting its theme (e.g., \textcolor{brown}{[CRITERION\_EXAMPLES]}). \\
2. Ensure that \textcolor{brown}{[CRITERION\_NAME]}s do not overlap in meaning, with each covering a single \textcolor{brown}{[CRITERION\_NAME]}. \\
3. Ensure that \textcolor{brown}{[CRITERION\_NAME]}s cover \textbf{ALL} provided relation types. \\
4. Provide a brief yet precise description (\textasciitilde 15 words) for each \textcolor{brown}{[CRITERION\_NAME]}.\\ 
5. Return the result in a valid JSON format as shown below. \\
\textbf{JSON Output Format: }\\
\verb+```+json\\
\{\\
  ``name of \textcolor{brown}{[CRITERION\_NAME]} 1'': ``description of \textcolor{brown}{[CRITERION\_NAME]} 1'', \\
  ``name of \textcolor{brown}{[CRITERION\_NAME]} 2'': ``description of \textcolor{brown}{[CRITERION\_NAME]} 2'',\\ 
  ...\\
\}\\
\verb+```+\\
\textbf{Relation Types and Descriptions: }\\
\verb+```+json\\
\textcolor{brown}{[RELATION\_WITH\_DESC]}\\
\verb+```+\\
\textbf{Output: }
\end{tcolorbox}

Next, we assign each relation to the most suitable tree nodes. 
In this template,
\textcolor{brown}{[CRITERION\_NAME]} denotes a textual criterion, \textcolor{brown}{[REL\_EXPLANATION]} refers to a predefined relation, \textcolor{brown}{[REL\_DESC]} is its description, and \textcolor{brown}{[CRITERION\_INSTANCES]} corresponds to the node names generated in the previous step.

\begin{tcolorbox}[
colback=gray!10, colframe=gray!90, 
title=Prompt Template for Relation Assignment, 
breakable]
\textbf{Task: }\\
You are tasked with analyzing the \textcolor{brown}{[CRITERION\_NAME]} of the relation label ``\textcolor{brown}{[REL\_NAME]}'' based on its description: ``\textcolor{brown}{[REL\_DESC]}''.\\
Determine which \textcolor{brown}{[CRITERION\_NAME]}(s) best align with ``\textcolor{brown}{[REL\_NAME]}''.\\
\textbf{Available \textcolor{brown}{[CRITERION\_NAME]}s: }\\
Below is a list of \textcolor{brown}{[CRITERION\_NAME]}s with their descriptions:\\
\verb+```+json\\
\textcolor{brown}{[CRITERION\_INSTANCES]}\\
\verb+```+\\
\textbf{Output Format: }\\
Provide your answer as a NON-EMPTY JSON array containing the \textcolor{brown}{[CRITERION\_NAME]}(s):\\
\verb+```+json\\
\hspace{0pt}[``\textcolor{brown}{[CRITERION\_NAME]} 1'', ``\textcolor{brown}{[CRITERION\_NAME]} 2'', ...]\\  
\verb+```+\\
\textbf{Output: }
\end{tcolorbox}

\begin{table*}[htbp]
\centering
\begin{tabular}{cccc}
\toprule
\textbf{Tree Construction Parameters} & \textbf{Value}        & \textbf{micro F1} & \textbf{binary F1} \\ \midrule
\multirow{3}{*}{Random Seed} & 42                      & 45.35 & 58.19 \\
                             & 666                     & 44.50 & 57.42 \\
                             & 1024                    & 44.55 & 57.98 \\ \midrule
\multirow{3}{*}{Criterion Set}        & (Domain, Entity Type) & 45.35             & 58.19              \\
                             & (Domain, Polarity)      & 45.10 & 57.08 \\
                             & (Entity Type, Polarity) & 44.14 & 57.36 \\ \bottomrule
\end{tabular}
\caption{Experiment results with varying tree construction parameters. }
\label{tab:robst}
\end{table*}

\subsection{API Cost of Tree Construction}

\begin{table}[!ht]
\centering
\begin{tabular}{ccc}
\toprule
\textbf{\#Input Tokens}  & \textbf{\#Output Tokens} & \textbf{Total} \\ \midrule
334,793          & 12,713      &    347,506    \\ \midrule
\textbf{Input Cost}  & \textbf{Output Cost} & \textbf{Total} \\ \midrule
\$0.8370          & \$0.1271      &    \$0.9641    \\ \bottomrule
\end{tabular}
\caption{API cost of hierarchical relation tree construction. }
\label{tab:lt-cost}
\end{table}

According to the cost analysis in Table \ref{tab:lt-cost}, constructing one relation tree with GPT-4o incurs a total API cost of \$0.9641. 
Since tree construction is performed only once for each relation schema, the overall overhead is 
negligible, underscoring the economic efficiency of our proposed pipeline.

\subsection{Statistics of Hierarchical Relation Tree}

We provide the detailed statistics of our hierarchical relation tree in Table~\ref{tab:stat-lt}. 
It is noted that there are more leaf nodes than predefined relations in the hierarchical relation tree. 
This occurs because some predefined relations inherently correspond to multiple high-level concepts, and thus may be assigned to more than one node.
For instance, it is reasonable for the relation ``\textit{significant person}'' to be contained by both intermediate nodes ``\textit{politics}'' and ``\textit{creative works}''.

\subsection{Robustness of Hierarchical Relation Tree}

Here, we construct hierarchical relation trees with varying random seeds and criterion sets. 
We first assess the tree edit similarities~\cite{tree-edit-dist} among trees generated with different random seeds, and then evaluate the overall performance of our model across all constructed trees. 
The results are reported in Table~\ref{tab:tree-sim} and Table~\ref{tab:robst}. 
While the relation trees exhibit some differences, the overall performance of our model remains consistently stable across different trees.

\subsection{Quality of Hierarchical Relation Tree}

\begin{figure*}[htbp]
    \centering
    \includegraphics[width=0.9\linewidth]{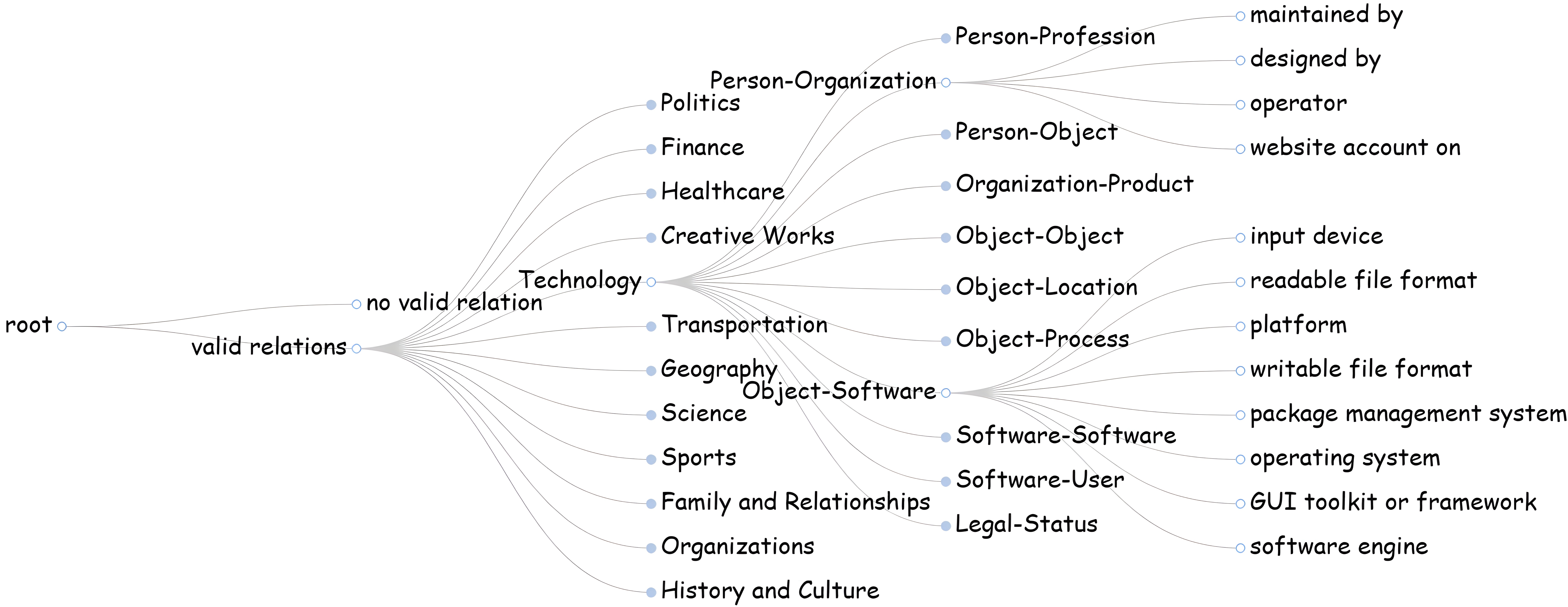}
    \caption{
        Visualization of a partial hierarchical relation tree. 
    }
    \label{fig:lt-vs}
\end{figure*}

To further assess the quality of our hierarchical relation tree, 
{
we provide both qualitative and quantitative evidence. 
}

{
\textbf{Qualitative Evidence.}
We conduct a case study to analyze the semantic coherence in intermediate nodes of the constructed tree.}
A partial view of the constructed tree is presented in Figure~\ref{fig:lt-vs}. 
Due to space limitations, only a representative subset of the full tree is shown.

\begin{table}[!t]
\centering
\begin{tabular}{cccc}
\toprule
\textbf{Seed} & 42    & 666   & 1024  \\ \midrule
42            & 100.0 & --    & --    \\
666           & 46.67 & 100.0 & --    \\ 
1024          & 47.82 & 47.72 & 100.0 \\ \bottomrule
\end{tabular}
\caption{The tree edit similarity (\%) matrix between constructed tree pairs. }
\label{tab:tree-sim}
\end{table}

\begin{table}[!t]
\centering
\begin{tabular}{cc}
\toprule
\textbf{Statistic}  & \textbf{Value} \\ \midrule
Tree Depth          & 5              \\
\#Node at Level 0   & 1              \\
\#Node at Level 1   & 2              \\
\#Node at Level 2   & 12             \\
\#Node at Level 3   & 136            \\
\#Node at Level 4   & 675            \\
\#Leaf Node         & 676            \\
\#Intermediate Node & 149            \\
\#Child Node (Avg.) & 5.50            \\ \bottomrule
\end{tabular}
\caption{The statistics of our hierarchical relation tree. }
\label{tab:stat-lt}
\end{table}

For the example in Figure~\ref{fig:lt-vs}, we adopt $\{$``\textit{domain}'', ``\textit{entity types}''$\}$ as the criterion set.
As shown in the partial tree structure, using the ``\textit{domain}'' criterion, the valid relations are first grouped into 12 high-level relation categories, such as ``\textit{politics}'', ``\textit{finance}'', and ``\textit{healthcare}''. 
Within each domain, we further organize the relations according to the ``\textit{entity types}'' criterion, resulting in multiple domain-specific subcategories. 
For instance, the domain ``\textit{technology}'' contains 11 entity types, such as ``\textit{object-software}'', ``\textit{software-user}'', and ``\textit{person-organization}''.
At the last level, each predefined relation (e.g., ``\textit{software engine}'' or ``\textit{GUI toolkit or framework}'') is linked to its corresponding 
tree nodes
(e.g. ``\textit{object-software}'').
By progressively narrowing the semantic scope from general concepts to fine-grained relations, the resulting tree forms a reasonable hierarchical structure that helps the LLM to conduct hierarchical classification. 

{
\textbf{Quantitative Evidence. }
To provide the explicit quantitative evidence, we follow \citet{hao-etal-2023-reasoning, wang2025speculative} and conduct an automated semantic evaluation using an advanced LLM GPT-4o. 
We iterate through all parent-child pairs in our tree and query GPT-4o with the prompt:
}
\begin{tcolorbox}[
colback=gray!10, colframe=gray!90, 
title=Prompt Template for Coherence Evaluation, 
breakable]
Is the relation node ``\textcolor{brown}{[CHILD\_NODE]}'' coherent with the node ``\textcolor{brown}{[PARENT\_NODE]}''? Output Yes/No only.
\end{tcolorbox}

{
We use the average probability of the ``Yes'' token output by GPT-4o as the \textit{coherence} score, and evaluate coherence for each pair of adjacent tree levels. 
}

\begin{table}[!t]
\centering
\begin{tabular}{cc}
\toprule
  & \textbf{Coherence (\%)} \\ \midrule
Level 1 $\rightarrow$ 2	   & 82.52              \\
Level 2 $\rightarrow$ 3	   & 80.27              \\
Level 3 $\rightarrow$ 4	   & 83.42              \\ \bottomrule
\end{tabular}
\caption{The semantic coherence scores for each pair of adjacent tree levels. }
\label{tab:coherence}
\end{table}

{
As shown in Table~\ref{tab:coherence}, every pair of adjacent tree levels achieves a high coherence score, demonstrating that our hierarchical relation tree is not only effective for downstream performance but also linguistically interpretable and meaningful, validating the semantic quality of intermediate nodes.
}

\section{Baselines}\label{ap:baseline}
We primarily compare our model with two categories of baselines: 

(1) \textit{Cross-Document RE Baselines}. 
\textbf{End-to-End}~\cite{yao-etal-2021-codred} is an Encoder-only model equipped with a selective attention mechanism to aggregate relation representations. 
\textbf{ECRIM}~\cite{wang-etal-2022-ecrim} introduces a cross-path entity relation attention mechanism to model the interaction among different text paths. 
\textbf{KXDocRE}~\cite{jain-etal-2024-kxdocre} enriches input text with domain knowledge to enhance relation representations.
\textbf{REIC}~\cite{na-etal-2024-reic} designs a reinforcement learning-based sentence selector to identify relation evidence.
\textbf{NEPD}~\cite{yue-etal-2024-nepd} integrates entity graph encoding with ECRIM and calibrates the prediction distribution. 

(2) \textit{LLM-based Hierarchical Text Classification (HTC) Baselines}. 
Since our model involves hierarchical classification during inference, we also reproduce several representative LLM-based HTC baselines for cross-document RE to provide a comprehensive comparison. 
\textbf{Rs-ICL}~\cite{chen-etal-2024-rsicl} uses a hierarchy-aware indexer to retrieve demonstrations for in-context learning in HTC. 
\textbf{DFS-L.}~\cite{yu2022constrained} and \textbf{BFS-L.}~\cite{huang2022exploring, jain-etal-2024-higen} train models to perform hierarchical classification by converting label hierarchies into sequences following depth-first and breadth-first search orders, respectively.

In addition to the above baselines, we also compare our model with a baseline referred to as \textbf{Vanilla}, which is fine-tuned to directly select the target relation from the full predefined relation set given the input instance. 

\section{Implementation Details}\label{ap:impl}

We choose LLaMA-3.1-8B-Instruct~\cite{grattafiori-2024-llama3} and GPT-4o as $\mathcal M_1$ and $\mathcal M_2$, respectively. 
During training, we adopt an AdamW optimizer~\cite{loshchilov-2019-adamw} with a learning rate of 5e-5 and a total batch size of 32.
Our model is trained for 6,400 steps using LoRA~\cite{hu2021loralowrankadaptationlarge} with $r$ = 64 and $\alpha$ = 128. 
All experiments are conducted on 4 NVIDIA A100 80G GPUs. 
To ensure fair comparisons, 
we adopt
the document-context filter of ECRIM~\cite{wang-etal-2022-ecrim} to preprocess all text paths. 

\section{The \textit{w/o} LTC Variant}\label{ap:vanilla-lt-build}

\begin{figure}[tbp]
    \centering
    \includegraphics[width=1\linewidth]{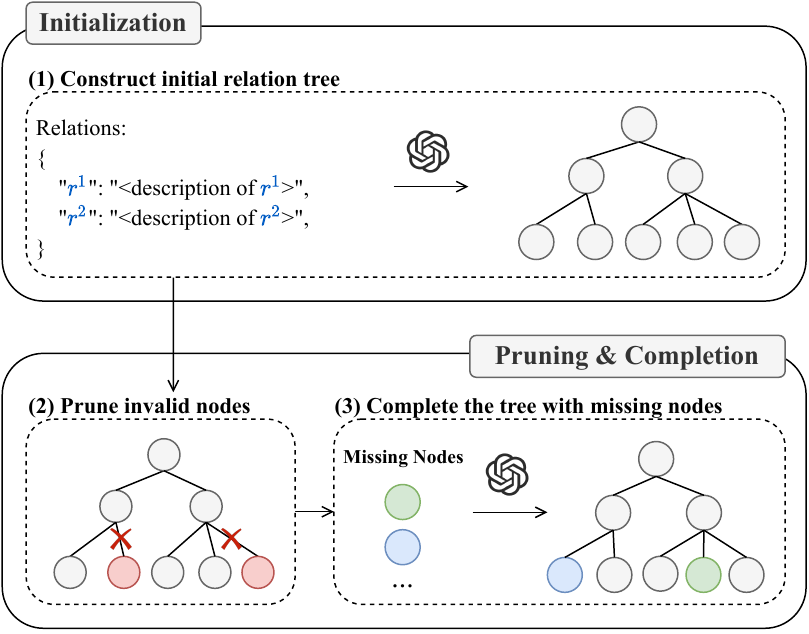}
    \caption{The tree construction pipeline for the \textit{w/o} LTC variant. }
    \label{fig:lt-old}
\end{figure}

Different from the level-wise pipeline of tree construction, in this variant, we first prompt the LLM $\mathcal M_2$ with predefined relations to directly generate a hierarchical relation tree in JSON format. 
Then, we ensure the completeness and validity of this tree step by step, as shown in Figure~\ref{fig:lt-old}. 
Concretely, we construct the hierarchical relation tree of the \textit{w/o}~LTC variant via the following steps. 

\textbf{Step 1:}
We begin by concatenating the names and descriptions of all predefined relations into a JSON string, and then input it into $\mathcal{M}_2$ to cluster and summarize the relations, producing an initial hierarchical relation tree. 

\textbf{Step 2:}
However, since LLMs are prone to hallucinations, the initial tree often contains invalid relations that are not involved in the predefined relations. 
For the invalid relations, we simply remove them from the relation tree. 

\textbf{Step 3:}
Besides, some predefined relations may be absent in the initial relation tree. 
For each missing relation, we prompt $\mathcal M_2$ to place it at a suitable position within the current relation tree. 

Following the above pipeline, we can easily obtain an LLM-generated hierarchical relation tree. 
Nevertheless, since the main tree structure is generated by $\mathcal M_2$ in a single run, the resulting trees often exhibit limited reasonableness.
This inherent limitation accounts for its worse performance compared with our level-wise pipeline.

\begin{table}[tbp]
\setlength{\tabcolsep}{0.43em}
\centering
\begin{tabular}{clccc}
\toprule
\textbf{Level} & \multicolumn{1}{c}{\textbf{Model}} & \textbf{\%CP} & \textbf{\%WP} & \textbf{\%SC} \\ \midrule
\multirow{2}{*}{1} & HCRE \textit{w/o} PtV & 34.64 & 0.00  & 65.36 \\
                   & HCRE (Ours)           & 72.67 & 0.00  & 27.33 \\ \midrule
\multirow{2}{*}{2} & HCRE \textit{w/o} PtV & 32.13 & 65.36 & 2.51  \\
                   & HCRE (Ours)           & 68.92 & 27.33 & 3.75  \\ \midrule
\multirow{2}{*}{3} & HCRE \textit{w/o} PtV & 31.31 & 67.87 & 0.82  \\
                   & HCRE (Ours)           & 64.86 & 31.08 & 4.06  \\ \midrule
\multirow{2}{*}{4} & HCRE \textit{w/o} PtV & 28.07 & 68.69 & 3.24  \\
                   & HCRE (Ours)           & 55.20 & 35.14 & 0.67  \\ \bottomrule
\end{tabular}
\caption{The error transfer details for HCRE \textit{w/o} PtV and HCRE. 
Note that \textbf{CP}, \textbf{WP}, and \textbf{SC} refer to ``Correct Prediction'', ``Wrong Parent'', and ``Sibling Confusion'', respectively. }
\label{tab:err-transfer}
\end{table}

\section{Error Types in Error Propagation}\label{ap:err-trans}

\begin{table*}[t]
\centering
\setlength{\tabcolsep}{0.2em}
\begin{tabular}{lcccc}
\toprule
\multicolumn{1}{c}{\multirow{2}{*}{\textbf{Model}}} & \multicolumn{2}{c}{\textbf{Closed}} & \multicolumn{2}{c}{\textbf{Open}} \\
\multicolumn{1}{c}{} & \textbf{bag-level micro F1} & \textbf{bag-level binary F1} & \textbf{bag-level micro F1} & \textbf{bag-level binary F1} \\ \midrule
RoBERTa+NEPD & 54.17 & 72.04 & 50.39 & 64.38 \\
Vanilla & 47.53 & 52.47 & 47.67 & 54.49 \\
HCRE (Ours) & \textbf{64.82} & \textbf{86.18} & \textbf{58.98} & \textbf{79.90} \\ \bottomrule
\end{tabular}
\caption{Bag-level evaluation results on CodRED under both closed and open settings. }
\label{tab:bag-agg}
\end{table*}

\begin{table*}[htbp]
\setlength{\tabcolsep}{0.48em}
\centering
\begin{tabular}{clccccccc} 
\toprule
  \multicolumn{1}{c}{\multirow{2}{*}{\textbf{Setting}}} & 
  \multicolumn{1}{c}{\multirow{2}{*}{\textbf{Model}}} &
  \multicolumn{4}{c}{\textbf{Dev}} &
  \multicolumn{2}{c}{\textbf{Test}} &
  \multicolumn{1}{c}{\multirow{2}{*}{\textbf{Avg.}}} \\ 
  &   
  &   
  \textbf{F1} &
  \textbf{AUC} &
  \textbf{P@500} &
  \textbf{P@1000} &
  \textbf{F1} &
  \textbf{AUC} & \\ \midrule
\multicolumn{1}{c}{\multirow{6}{*}{\textbf{Closed}}}
& End-to-End \cite{yao-etal-2021-codred} & 51.26 & 47.94 & 62.80 & 51.00 & 51.02 & 47.46 & 49.42 \\
& ECRIM \cite{wang-etal-2022-ecrim} & 61.12 & 60.91 & 78.89 & 60.17 & 62.48 & 60.67 & 61.30 \\
& KXDocRE \cite{jain-etal-2024-kxdocre} & {64.97} & 64.30 & -- & -- & {66.30} & 65.55 & {65.28} \\
& REIC \cite{na-etal-2024-reic} & 63.47 & {66.41} & {80.20} & 63.50 & 65.02 & {65.88} & 65.20 \\
& NEPD \cite{yue-etal-2024-nepd} & 63.63 & 65.01 & 77.84 & {64.03} & 64.41 & \textbf{66.23} & 64.82 \\ 
\cmidrule{2-9}
& HCRE (Ours) & \textbf{65.36} & \textbf{67.06} & \textbf{80.40} & \textbf{64.60} & \textbf{66.91} & 64.40 & \textbf{65.93} \\ \midrule
\multicolumn{1}{c}{\multirow{4}{*}{\textbf{Open}}}
& End-to-End \cite{yao-etal-2021-codred} & 47.23 & 40.86 & 59.00 & 46.30 & 45.06 & 39.05 & 43.05 \\
& KXDocRE \cite{jain-etal-2024-kxdocre} & {56.70} & {55.20} & -- & -- & {57.93} & \textbf{57.12} & {56.74} \\
& NEPD \cite{yue-etal-2024-nepd} & 54.29 & 54.92 & {68.66} & {53.84} & 56.68 & 55.87 & 55.44 \\ \cmidrule{2-9}
& HCRE (Ours) & \textbf{58.15} & \textbf{55.79} & \textbf{70.60} & \textbf{57.50} & \textbf{60.85} & {56.34} & \textbf{57.78} \\ \bottomrule
\end{tabular}
\caption{
    Experiment results on conventional metrics under both closed and open settings. 
    \textbf{F1} denotes maximum F1, and \textbf{Avg.} scores are computed based on F1 and AUC only.
    Baseline results are taken from their respective original papers. 
}
\label{tab:conv-metric}
\end{table*}

\begin{table*}[htbp]
\centering
\setlength{\tabcolsep}{0.8em}
\begin{tabular}{clcccc}
\toprule
\multirow{2}{*}{\textbf{Backbone}} &
  \multicolumn{1}{c}{\multirow{2}{*}{\textbf{Model}}} &
  \multicolumn{2}{c}{\textbf{Closed}} &
  \multicolumn{2}{c}{\textbf{Open}} \\
 &
  \multicolumn{1}{c}{} &
  \textbf{micro F1} &
  \textbf{binary F1} &
  \textbf{micro F1} &
  \textbf{binary F1} \\ \midrule
\multirow{2}{*}{Qwen2.5-0.5B-Instruct} & Vanilla     & 18.63          & 26.41          & 7.00           & 10.15          \\
                                       & HCRE (Ours) & \textbf{25.83} & \textbf{44.10} & \textbf{21.00} & \textbf{37.63} \\ \midrule
\multirow{2}{*}{Qwen2.5-7B-Instruct}   & Vanilla     & 36.10          & 41.16          & 14.20          & 16.59          \\
                                       & HCRE (Ours) & \textbf{37.29} & \textbf{52.35} & \textbf{34.40} & \textbf{53.43} \\ \midrule
\multirow{2}{*}{Gemma2-9B-Instruct}    & Vanilla     & 37.34            & 41.77            & 14.53            & 17.04            \\
                                       & HCRE (Ours) & \textbf{41.74}   & \textbf{53.39}   & \textbf{28.94}   & \textbf{38.75}   \\ \bottomrule
\end{tabular}
\caption{Experiment results on CodRED with different backbones. }
\label{tab:backbone}
\end{table*}

To gain deeper insights into how errors propagate through the hierarchical classification process, we categorize error cases into two primary types: ``Wrong Parent'' and ``Sibling Confusion''.
The detailed error transfer statistics are summarized in Table~\ref{tab:err-transfer}.
As shown in the table, PtV greatly reduces errors of the ``Wrong Parent'' type. 
Since an incorrect parent node almost guarantees an incorrect prediction, reducing this type of error effectively alleviates error propagation during hierarchical classification.

{
\section{Bag-level Evaluation}\label{ap:bag}
To further align the evaluation protocals with prior works, we perform bag-level evaluation by applying a simple majority voting mechanism to aggregate the path-level results into bag-level (entity-pair-level) predictions. 
Specifically:
\begin{enumerate}
    \item The LLM independently generates a relation prediction for every text path within a bag. The confidence scores of predicted relations are also recorded for tie-breaking.
    \item For each bag, the most frequent positive relation is selected as the bag-level prediction.
    \item If multiple relations are tied for the highest frequency, we choose the relation with the highest confidence as the bag-level prediction.
    \item If all paths within the bag are predicted as NA, the bag-level prediction is NA.
\end{enumerate}
Notably, the bag-level evaluation is essentially another form of path-level evaluation, requiring no additional model training or inference. 
Therefore, the two evaluation protocols are fundamentally consistent.
}

{
As shown in Table~\ref{tab:bag-agg}, HCRE still significantly outperforms the second-best baseline (RoBERTa+NEPD) and Vanilla at bag-level evaluation. 
This is consistent with our path-level evaluation, demonstrating that our evaluation ensures fairness and practical usability.
}

\section{Experiments on Conventional Metrics}\label{ap:conv-metric}

Following prior work in cross-document RE~\cite{yao-etal-2021-codred, wang-etal-2022-ecrim, yue-etal-2024-nepd}, we evaluate our model on CodRED using conventional metrics, including P@K, AUC, and maximum F1. 
We compare HCRE with several representative cross-document RE models: 
End-to-End~\cite{yao-etal-2021-codred}, 
ECRIM~\cite{wang-etal-2022-ecrim}, 
KXDocRE~\cite{jain-etal-2024-kxdocre}, 
REIC~\cite{na-etal-2024-reic}, 
and 
NEPD~\cite{yue-etal-2024-nepd}. 
Details of these baselines are provided in Appendix~\ref{ap:baseline}.  
For HCRE, to compute the conventional score-based metrics, we traverse the relation tree, obtain the score for each node and aggregate them to produce the relation score distribution.

As shown in Table~\ref{tab:conv-metric}, HCRE consistently outperforms all baselines across the conventional metrics under both closed and open settings. 
Furthermore, we submit the test results to the official competition leaderboard, where our model obtains 66.91 and 60.85 F1 points under the closed and open settings, respectively.

\begin{table}[!t]
\centering
\begin{tabular}{lcccc}
\toprule
\multicolumn{1}{c}{\multirow{2}{*}{\textbf{Model}}} & \multicolumn{2}{c}{\textbf{Dev}} & \multicolumn{2}{c}{\textbf{Test}} \\ 
\multicolumn{1}{c}{}                                & \textbf{F1}   & \textbf{Ign F1}   & \textbf{F1}   & \textbf{Ign F1}   \\ \midrule
\multicolumn{5}{c}{\textit{\textbf{LLM-based Doc-level RE Baselines}}} \\ \midrule
AutoRE & 50.37 & 48.79 & 50.35 & 48.58 \\
EP-RSR & 56.17 & 53.74 & 56.47 & 53.22 \\ \midrule
\multicolumn{5}{c}{\textit{\textbf{Ours}}} \\ \midrule
Vanilla     & 61.30 & 60.21 & 60.77 & 59.63 \\ 
HCRE        & \textbf{62.80} & \textbf{61.22} & \textbf{61.57} & \textbf{60.18} \\ \bottomrule
\end{tabular}
\caption{Experiment results on DocRED. }
\label{tab:docred}
\end{table}

\section{Experiments on Different Backbones}\label{ap:backbone}

Due to the widespread use of LLaMA-3.1-8B-Instruct in the research community, we use it as our primary backbone for experiments.
To examine the generality of HCRE across backbone architectures and model scales, we further report the results of experiments using Qwen2.5-0.5B-Instruct, Qwen2.5-7B-Instruct~\cite{qwen2025qwen25technicalreport}, and Gemma2-9B-Instruct~\cite{gemmateam2024gemma2improvingopen}. 
As shown in Table~\ref{tab:backbone}, HCRE consistently yields performance gain, suggesting our PtV inference strategy is robust to both backbone architecture and model scale.

\section{Cross-Dataset Generalization of HCRE}\label{ap:docred}

To further validate the generalization of HCRE across different datasets, we examine the performance of HCRE on DocRED~\cite{yao-etal-2019-docred}, a popular document-level RE benchmark. 

Following the prior studies in document-level RE~(\citealp{zhang-etal-2022-docre1}; \citealp{ijcai2023p586}; \citealp{Zhang_Su_Min_Miao_Hu_Fu_Shi_Chen_2023}; \citealp{zhang-etal-2025-eprsr}), we adopt F1 and Ign~F1 as our evaluation metrics. 
We compare HCRE with the Vanilla baseline and two representative LLM-based document-level RE models: 
1) \textbf{AutoRE}~\cite{xue-etal-2024-autore} proposes a LLM-based Relation-Head-Facts paradigm that enables the LLM to extract relations without the need to perceive the full predefined relation set.
2) \textbf{EP-RSR}~\cite{zhang-etal-2025-eprsr} introduces an Entity-Pair–Relation–Fact paradigm and enhances the relevance between candidate relations and target entities via an entity pair-level relation filtering method.
As shown in Table~\ref{tab:docred}, HCRE consistently surpasses these baselines on both development and test sets across all metrics, demonstrating strong cross-dataset generalization.
\begin{algorithm*}[!ht]
   \caption{Prediction-then-Verification Inference Strategy}
   \label{alg:ptv}
\begin{algorithmic}
\STATE {\bfseries Input:} LLM $\mathcal M_1$,\ hierarchical relation tree $\mathcal T$, context $c$, head entity $e_h$, tail entity $e_t$, current level $l$, node predicted at ($l$-$1$)-th level $\hat r_{l-1}$, and maximum PtV round $M$
\STATE {\bfseries Output:} Final relation $\hat r_l$

\STATE $\mathcal R_l \gets \mathcal T\text{.children\_of}(\hat r_{l-1})$
\STATE $t \gets 0$  \textit{// Number of PtV rounds}

\WHILE{True}


    \STATE \textit{// Step 1: Prediction step}
    \STATE $\hat r_\text{1st}, \hat r_\text{2nd} \gets \mathcal M_1(c, e_h, e_t, \mathcal R_l)$
    
    \STATE \textit{// Step 2: Verification step}
    \STATE \textit{// Sub-step 2.1: Replace nodes with their respective children}
    \STATE $\mathcal R_l^{v_1} \gets \mathcal R_l\text{.replace}(\hat r_\text{1st}, \mathcal T\text{.children\_of}(\hat r_\text{1st}))$
    \STATE $\mathcal R_l^{v_2} \gets \mathcal R_l\text{.replace}(\hat r_\text{2nd}, \mathcal T\text{.children\_of}(\hat r_\text{2nd}))$
    \STATE $\mathcal R_l^{v_3} \gets \mathcal R_l\text{.replace}(\hat r_\text{1st}, \mathcal T\text{.children\_of}(\hat r_\text{1st}))\text{.replace}(\hat r_\text{2nd}, \mathcal T\text{.children\_of}(\hat r_\text{2nd}))$
    
    \STATE \textit{// Sub-step 2.2: Verify }$\hat r_\text{1st}$
    \STATE $\hat r^{v_1},\_ \gets \mathcal M_1(c, e_h, e_t, \mathcal R_l^{v_1})$
    \STATE $\hat r^{v_2},\_ \gets \mathcal M_1(c, e_h, e_t, \mathcal R_l^{v_2})$
    \STATE $\hat r^{v_3},\_ \gets \mathcal M_1(c, e_h, e_t, \mathcal R_l^{v_3})$
    \IF{$\mathbbm{1}_{\hat r^{v_1}\in \mathcal T\text{.children\_of}(\hat r_\text{1st})} + \mathbbm{1}_{\hat r^{v_2} = \hat r_\text{1st}} + \mathbbm{1}_{\hat r^{v_3}\in \mathcal T\text{.children\_of}(\hat r_\text{1st})} \geq 2$}
        \STATE $\hat r_l \gets \hat r_\text{1st}$
        \STATE \textbf{break}
    \ELSE
        \STATE $\mathcal R_l\text{.remove}(\hat r_\text{1st})$
    \ENDIF
    
    \STATE \textit{// Exceeds max round limitation}
    \STATE $t \gets t + 1$
    \IF{$t > M$}
        \STATE $\hat r_l \gets \hat r_\text{1st}$
        \STATE \textbf{break}
    \ENDIF
\ENDWHILE
\RETURN $\hat r_l$
\end{algorithmic}
\end{algorithm*}

\section{More Details about PtV}\label{ap:ptv-more}

\begin{table}[!t]
\setlength{\tabcolsep}{0.16em}
\centering
\begin{tabular}{lccc}
\toprule
\multicolumn{1}{c}{\textbf{Model}} & \textbf{Input Tok.} & \textbf{\#LLM Calls} & \multicolumn{1}{l}{\textbf{Latency}} \\ \midrule
Vanilla                             & 1,499.76             & 40,740               & 0.21s                                \\
HCRE (Ours)                         & 575.75              & 152,663              & 0.29s                                \\ \bottomrule
\end{tabular}
\caption{Comparison of computation efficiency between Vanilla and HCRE. \textbf{Input Len.} refers to the average input tokens. }
\label{tab:comp-comp}
\end{table}

\begin{figure}
    \centering
    \includegraphics[width=1\linewidth]{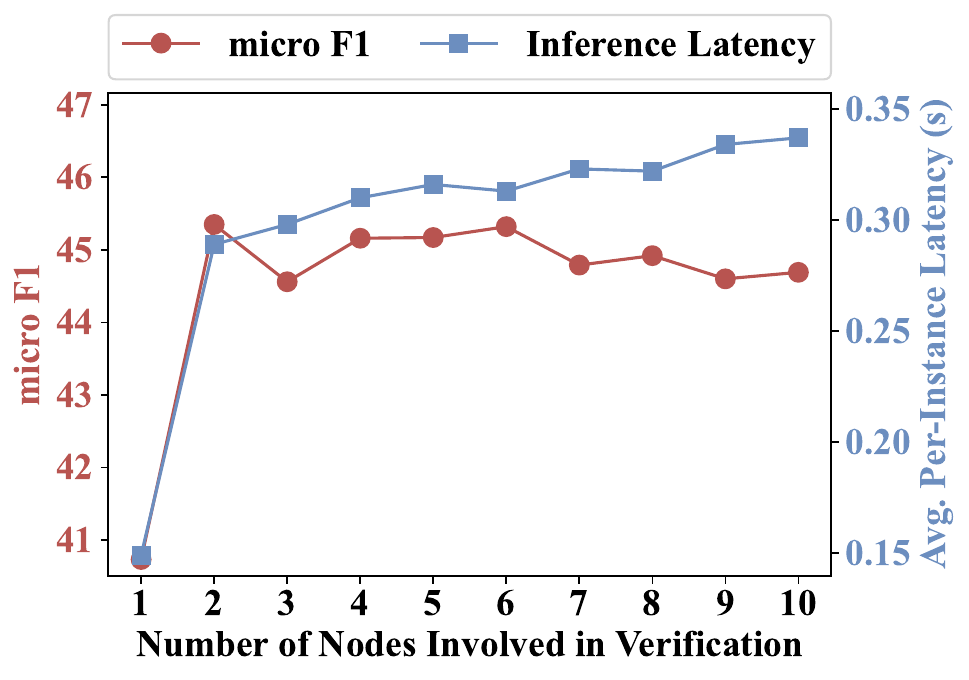}
    \caption{The micro F1 and per-instance inference latency using PtV with varying numbers of verification nodes. }
    \label{fig:top_v}
\end{figure}

To further demonstrate our PtV inference strategy, we further describe its detailed procedure, analyze its computation efficiency, 
{
and conduct ablation experiments on the number of nodes involved in verification
} in this section. 

\subsection{Detailed Procedure}

To describe the PtV inference strategy precisely, we present the complete PtV process in Algorithm~\ref{alg:ptv}. 

\subsection{Computation Efficiency}\label{ap:comp-comp}

To analyze the computational efficiency of the PtV inference strategy, we compare the average per-instance input tokens, the number of LLM calls, and the average per-instance latency of Vanilla and HCRE on an NVIDIA A100 80G GPU.
The statistics are presented in Table~\ref{tab:comp-comp}. 
Although HCRE triggers more LLM calls, its hierarchical classification paradigm substantially reduces the number of options during each inference, reducing the average number of input tokens by 61.6\% (from 1,499.76 to 575.75). 
Given the $\mathcal O(N^2)$ computation complexity of Transformer~\cite{vaswani2023attentionneed}, this reduction accelerates the prefilling phase of LLM inference.
Consequently, HCRE achieves latency on par with Vanilla.

{
\subsection{Extending Verification to Top-\texorpdfstring{$k$}{k} Nodes}
To analyze the impact of the number of nodes involved in verification, we conduct an ablation experiment that extends verification to top-$k$ nodes, with $k$ ranging from 1 to 10.
}

{
As shown in Figure~\ref{fig:top_v}, the top-2 setting achieves a significant improvement over top-1, indicating that considering more than one candidate helps mitigate error propagation more effectively. 
However, further increasing $k$ beyond 2 does not bring more performance gain while adding computational overhead.
Therefore, our design choice maximizes the performance-efficiency trade-off.
}

\end{document}